\pdfoutput=1
\documentclass[11pt]{article}

\usepackage{naacl2021}

\usepackage{times}
\usepackage{latexsym}

\usepackage[T1]{fontenc}
\usepackage[utf8]{inputenc}

\usepackage{microtype}
\usepackage{amsthm}
\usepackage{algorithm}
\usepackage{helvet}
\usepackage{courier}
\usepackage{latexsym}
\usepackage{textcomp}
\usepackage{enumerate}
\usepackage{tabularx}
\usepackage{multirow}
\usepackage{adjustbox}
\usepackage{array}
\usepackage{graphics}
\usepackage{graphicx}
\usepackage{subfig}
\usepackage{color}
\usepackage{amsmath}
\usepackage{makecell}
\usepackage{indentfirst}
\usepackage{multirow}
\usepackage{amssymb}
\usepackage{algpseudocode}
\usepackage{comment}
\usepackage{mathrsfs}
\usepackage{extarrows}
\usepackage{booktabs}
\usepackage{setspace}

\title{Mask Attention Networks: Rethinking and Strengthen Transformer}

\author{Zhihao Fan$^{1}$\thanks{~~Work is done during internship at Microsoft Research Asia.}, Yeyun Gong$^{2}$, Dayiheng Liu$^{3}$, Zhongyu Wei$^{1,6}$\thanks{~~Corresponding author.}, Siyuan Wang$^{1}$, \\
\textbf{Jian Jiao}$^{4}$\textbf{, Nan Duan}$^{2}$\textbf{, Ruofei Zhang}$^{4}$\textbf{, Xuanjing Huang}$^{5}$\\
$^{1}$School of Data Science, Fudan University, China\\
$^{2}$Microsoft Research Asia, $^{3}$DAMO Academy, $^{4}$Microsoft \\
$^{5}$School of Computer Science, Fudan University, China\\
$^{6}$Research Institute of Intelligent and Complex Systems, Fudan University, China\\
\{fanzh18,zywei,wangsy18,xjhuang\}@fudan.edu.cn, liudayiheng.ldyh@alibaba-inc.com\\
\{yegong,Jian.Jiao,nanduan,bzhang\}@microsoft.com
}

\begin{document}

\maketitle
\begin{abstract}
Transformer is an attention-based neural network, which consists of two sublayers, namely, Self-Attention Network (SAN) and Feed-Forward Network (FFN). Existing research explores to enhance the two sublayers separately to improve the capability of Transformer for text representation. In this paper, we present a novel understanding of SAN and FFN as Mask Attention Networks (MANs) and show that they are two special cases of MANs with static mask matrices. However, their static mask matrices limit the capability for localness modeling in text representation learning. We therefore introduce a new layer named dynamic mask attention network (DMAN) with a learnable mask matrix which is able to model localness adaptively. To incorporate advantages of DMAN, SAN, and FFN, we propose a sequential layered structure to combine the three types of layers. Extensive experiments on various tasks, including neural machine translation and text summarization demonstrate that our model outperforms the original Transformer.
\end{abstract}

%\footnotetext[$*$]{Work is done during internship at Microsoft Research Asia.}
%\footnotetext[2]{Corresponding author.}

\section{Introduction}
Recently, Transformer~\cite{vaswani2017attention} has been widely applied in various natural language processing tasks, such as neural machine translation~\cite{vaswani2017attention} and text summarization~\cite{zhang2019pretraining}. To further improve the performance of the text representation, Transformer-based variants have attracted a lot of attention~\cite{lu2019understanding,sukhbaatar2019adaptive,sukhbaatar2019augmenting,bugliarello2019improving,ma2019monotonic}. 

Each building block of Transformer has two sublayers: Self-Attention Network (SAN) and Feed-Forward Network (FFN).
~\citet{shaw2018self} presents an extension to SAN which incorporates the relative positional information for the sequence.~\citet{sukhbaatar2019adaptive} proposes attention span to control the maximum context size used in SAN and scales Transformer to long-range $(\sim8192$ tokens) language modeling. Recently, some works targeting on FFN have been proposed.~\citet{lu2019understanding} gives a new understanding of Transformer from a multi-particle dynamic system point of view and designs a macaron architecture following Strang-Marchuk splitting scheme.~\citet{sukhbaatar2019augmenting} regards the FFN as the persistent memory in SAN to augment SAN. These works focus on enhancing SAN or FFN, but neglect the inner relationship between SAN and FFN that hinders further improvement. 

\begin{figure}[t]
    \centering
    \includegraphics[width = 3.0in]{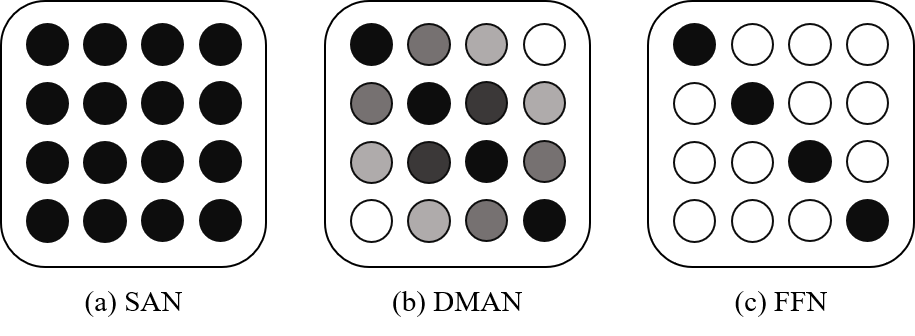}
	\caption{The mask matrices of (a) SAN, (b) DMAN and (c) FFN in Mask Attention Networks. Color that fades from black to white means the values in mask matrices decrease from 1 to 0.}\label{mask.matrix}
\end{figure}

% In this work, we present a more systematic analysis for both SAN and FFN to reveal their connections. We introduce \textit{Mask Attention Networks (MANs)}, in which each network has a mask matrix that element-wise multiplies a key-query attention matrix. We show that SAN and FFN are two particular cases in MANs with static mask matrices. The mask matrix of SAN is an all-ones matrix, while that of FFN is an identity matrix, which is shown as (a) and (c) in Figure~\ref{mask.matrix}. Since the mask matrix of SAN has no restriction on relationship modeling with other tokens, SAN is expert in long-range dependency modeling. In contrast, mask of FFN disables it to perceive the information of other tokens and forces it into self-evolution. We believe that these two specialties endowed by two mask matrices make the success of Transformer in text representation. However, the static mask matrices of SAN and FFN only allow them to model sequence on two special scales, respectively, and limit their capability in localness modeling. 

In this work, we present a more systematic analysis for both SAN and FFN to reveal their connections. We introduce \textbf{Mask Attention Networks}(MANs), in which each network has a mask matrix that element-wise multiplies a key-query attention matrix. We show that SAN and FFN are two special cases in MANs with static mask matrices. The mask matrix of SAN is an all-ones matrix, while that of FFN is an identity matrix, which is shown as (a) and (c) in Figure~\ref{mask.matrix}. Since the mask matrix of SAN has no restriction on relationship modeling with other tokens, SAN is expert in long-range dependency modeling and capture the global semantics. In contrast, mask of FFN disables it to perceive the information of other tokens and forces it into self-evolution. We believe that these two specialties endowed by two mask matrices make the success of Transformer in text representation. 

Although positive results of Transformer have been reported, recent works~\cite{shaw2018self,yang2018modeling,guo2019gaussian} have shown that modeling localness would further improve the performance through experiments. We argue that deficiency of Transformer in local structure modeling is caused by the attention computation with static mask matrix. 
In the framework of MANs, we find a problem that irrelevant tokens with overlapping neighbors incorrectly attend to each other with relatively large attention scores. For example ``a black dog jump to catch the frisbee'', though ``catch'' and ``black'' are neither relevant nor neighbors, for the reason that both of them are highly related to their common neighbor ``dog'' in attention, we demonstrate that the attention score from ``catch'' to ``black'' would be large, which also decreases the attention score from ``catch'' to ``frisbee''. The issue in self-attention not only introduces noise to the semantic modeling, but also mislead query tokens to overlook these neighbor tokens. This reveals that self-attention is insufficient in localness modeling and inspires us to mask tokens that not appear in neighborhood. 
% ~\citet{guo2019gaussian} and~\citet{yang2018modeling} utilize gaussian bias to model the scope of the locality. From the pespective of DMANs, we propose a flexible and general solution

To strengthen Transformer in localness modeling with better keeping the advantage of SAN and FFN, we propose a Dynamic Mask Attention Network (DMAN) as shown in Figure~\ref{mask.matrix}(b), which originates from MANs. Observations reveal that tokens have different ranges of neighbors, for example, that of ``dog'', which is also connected with ``frisbee'', is larger than ``black'' and ``catch''.
Instead of being static that determined in advance, the mask matrix of DMAN is dependent on the query context and relative distance. 
In DMAN, the tokens in a specific neighborhood are able to receive more attention beyond the normal self-attention mechanism. The dynamic endows DMAN with text representation in different scales, and we validate the superiority through experiments. In Transformer~\cite{vaswani2017attention}, SAN and FFN cooperate in a sequential layered structure SAN$\to$FFN. Considering SAN, FFN, and DMAN all belong to MANs and have different advantages in text representation, instead of directly replacing SAN in previous works~\cite{shaw2018self,yang2018modeling,guo2019gaussian}, we propose to incorporate them with the architecture DMAN$\to$SAN$\to$ FFN.

The main contributions of this work are threefold:
\begin{itemize}
%\item We introduce \textit{Mask Attention Network Family}and reformulate SAN and FFN layers from a mask matrix perspective. Though the analysis, we point out the mask matrices used in SAN and FFN layers are two extreme cases in. We analyze the advantages of SAN and FFN in text representation and prove that they are insufficient for localness modeling.
\item We introduce Mask Attention Networks and reformulate SAN and FFN to point out that they are two special cases with static mask in MANs. We analyze the advantages of SAN and FFN in text representation learning and demonstrate that they are insufficient for localness modeling.
%\item We propose a novel dynamic mask matrix to capture the general situation in the attention layer. Based on the dynamic mask, we propose Dynamic Mask Attention Network (DMAN) as the to more effectively model localness. We investigate the different collaboration methods of SAN, FFN and DMAN and verify the validity of our proposed architecture DMAN$\to$SAN$\to$FFN in experiments.
\item Inspired by the different specialities of SAN and FFN, we propose Dynamic Mask Attention Network (DMAN) to model localness more effectively. We investigate the different collaboration methods of SAN, FFN, and DMAN, and propose a sequential layered structure DMAN$\to$SAN$\to$FFN.
\item We conduct experiments on machine translation and abstract summarization. Experimental results show that our method outperforms original Transformer. We also perform ablation study to verify the effectiveness of different modules of our proposed model.
\end{itemize}

\section{Model}
%We propose Dynamic Mask Attention Network based on Transformer~\cite{vaswani2017attention}.
In \S~\ref{transformer-layer}, we review the Transformer architecture. We introduce Mask Attention Networks and reformulate SAN and FFN to point out they are two special cases in \S~\ref{mask-attention-network-family}, and analyze their deficiency in localness modeling in \S~\ref{deficiency-of-san-and-ffn-in-localness-modeling}. Then, in \S~\ref{dynamic-mask-self-attention-network}, we describe Dynamic Mask Attention Network (DMAN) in detail. At last, in \S~\ref{collaboration-of-mask-attention-network-family}, we discuss the collaboration of DMAN, SAN and FFN.

\subsection{Transformer}
\label{transformer-layer}
Transformer has two sublayers: Self-Attention Network (SAN) and Feed-Forward Network (FFN).

As discussed in~\citet{vaswani2017attention}, an attention function maps a query and a set of key-value pairs to an output shown in Equation~\ref{self-attention}.
\begin{equation}
    %\begin{small}
    \begin{gathered}
        \mathcal{A}(Q,K,V) = \mathcal{S}(Q,K)V \\ \mathcal{S}(Q,K)=\Bigg[\frac{\text{exp}\big(Q_{i}K_{j}^{T}/\sqrt{d_{k}}\big)}{\sum_{k}\text{exp}\big(Q_{i}K_{k}^{T}/\sqrt{d_{k}}\big)}\Bigg]
    \end{gathered}
    \label{self-attention}
    %\end{small}
\end{equation}
%Softmax\big(\frac{QK^T}{\sqrt{d_{k}}}\big) 
where the queries $Q$, keys $K$ and values $V\in\mathbb{R}^{T\times d_{k}}$ are all matrices. %The input consists of queries and keys of dimension $d_k$.

{\normalsize \setlength{\lineskip}{2em} SAN produces representations by applying attention function to each pair of tokens from the input sequence. It is beneficial to capture different contextual features with multiple individual attention functions. Given a text representation sequence $H^{l}\in \mathbb{R}^{T\times d}$. } in the $l$-the layer.%
\begin{equation}
    %\begin{small}
    \begin{gathered}
    H^{l}=\big[A^{1},\cdots,A^{I}\big] W_H \\
    A^{i}=\mathcal{A}\big(H^{l}W^{i}_{Q},H^{l}W^{i}_{K},H^{l}W^{i}_{V}\big) 
\end{gathered}
    %\end{small}
\end{equation}
where $\{W^{i}_{Q},W^i_K,W^i_V\}\in R^{d\times d_{k}}$ are trainable parameters, $i$ denotes the attention head and $d$ is the hidden size.

In FFN, the computation of each $h^{l}_{t}$ in $H^{l}$ is independent of others.
%The second element of a transformer layer is a fully connected feed-forward layer. This sublayer is applied to each position t in the input sequence independently, and 
It consists of two affine transformations with a pointwise non-linear function:
\begin{equation}
    %\begin{small}
        \begin{gathered}
        H^{l+1} = \text{ReLU}\big(H^{l}W_1\big)W_2
        \end{gathered}
    %\end{small}
\end{equation}
%where $\mathcal{F}(x) = \mathop{\max}(0,x)$ is the ReLU activation function, 
where $W_{1}$ and $W_{2}$ are matrices of dimension $d\times d_f$ and $d_f\times d$, respectively. Typically, $d_f$ is set to be 4 times larger than $d$.

\subsection{Mask Attention Networks}
\label{mask-attention-network-family}
On the basis of attention function in Equation~\ref{self-attention}, we define a new mask attention function:
\begin{equation}
    %\begin{small}
        \begin{gathered}
            \mathcal{A}_{M}(Q,K,V) =\
            {S}_{M}(Q,K)V \\
            \mathcal{S}_{M}(Q,K)=\Bigg[\frac{M_{i,j}\text{exp}\big(Q_{i}K_{j}^{T}/\sqrt{d_{k}}\big)}{\sum_{k}M_{i,k}\text{exp}\big(Q_{i}K_{k}^{T}/\sqrt{d_{k}}\big)}\Bigg]
        \end{gathered}
    %\end{small}
    \label{mask_attention_function_1}
\end{equation}
where $M\in\mathbb{R}^{T\times T},M_{i,j}\in[0,1]$ is a mask matrix and can be static or dynamic. Intuitively, the value in each position of $M$ can be viewed as the color shade in Figure~\ref{mask.matrix}.

With the knowledge of mask attention function, we introduce \textbf{Mask Attention Networks}(MANs), in which each network can be written as Equation~\ref{mask-attention}.
\begin{equation}
    %\begin{small}
        \begin{gathered}
            H^{l+1}=\mathcal{F}\big(\big[A^{1}_{M^{1}},\cdots,A^{I}_{M^{I}}\big]\big)W_{H} \\
            A^{i}_{M^{i}}=\mathcal{A}_{M^i}\big(H^{l}W^{i}_{Q},H^{l}W^{i}_{K},H^{l}W^{i}_{V}\big) \label{mask-attention}
        \end{gathered}
    %\end{small}
\end{equation}
where $\mathcal{F}$ is the activation function, $M^i$ is the mask matrix for the $i$-th attention head.

Next, we show that SAN and FFN both belong to the Mask Attention Networks.

For SAN, let $M=[1]\in\mathbb{R}^{T\times T}$ be an all-ones matrix and $\mathcal{F}=\mathcal{F}_{id}$ be the identity function, its mask attention function would be formalized:
\begin{equation}
    %\begin{small}
        \begin{gathered}
            \mathcal{S}_{[1]}(Q,K)=\Bigg[\frac{1\cdot\text{exp}\big(Q_{i}K_{j}^{T}/\sqrt{d_{k}}\big)}{\sum_{k}\text{exp}\big(Q_{i}K_{k}^{T}/\sqrt{d_{k}}\big)}\Bigg]=\mathcal{S}(Q,K) \\
            \mathcal{A}_{[1]}(Q,K,V)=\mathcal{S}_{[1]}(Q,K)V=\mathcal{A}(Q,K,V)
        \end{gathered}
    %\end{small}
\end{equation}

%&=\Bigg[\frac{exp\big(Q_{i}K_{j}^{T}\big)}{\sum_{k} exp\big(Q_{i}K_{k}^{T}\big)}\Bigg] \nonumber \\
Then, the MAN degenerates into SAN.
\begin{equation}
    %\begin{small}
        \begin{aligned}
            H^{l+1}&=\mathcal{F}_{id}\Big(\big[A^{1}_{[1]},\cdots,A^{h}_{[1]}\big]\Big)W_{H} \\
            &=\big[A^{1},\cdots,A^{h}\big]W_{H} 
        \end{aligned}
    %\end{small}
\end{equation}
%&=\big[A^{1}_M,\cdots,A^{h}_M\big]W_{H} \nonumber \\

For FFN, let $M=\mathbb{I}\in\mathbb{R}^{T\times T}$ be the identity matrix, $\mathcal{F}=\text{ReLU}$ and head number $I=1$. 
\begin{equation}
    %\begin{small}
        \begin{gathered}
        \mathcal{S}_{\mathbb{I}}(Q,K)=\Bigg[\frac{1_i(j)\cdot \text{exp}\big(Q_{i}K_{j}^{T}/\sqrt{d_k}\big)}{\sum_{k}1_i(k)\cdot \text{exp}\big(Q_{i}K_{k}^{T}/\sqrt{d_{k}}\big)}\Bigg]=\mathbb{I} \\
        \mathcal{A}_{\mathbb{I}}(Q,K,V) =\mathcal{S}_{\mathbb{I}}(Q,K)V =\mathbb{I}V =V
        \end{gathered}
    %\end{small}
\end{equation}
where $1_{i}(x)$ is an indicator function that equal to 1 if $x=i$, otherwise 0. 

%&=\Bigg[\frac{exp\big(Q_{i}K_{j}^{T}\big)}{\sum_{k} exp\big(Q_{i}K_{k}^{T}\big)}\Bigg] \nonumber \\
The MAN degenerates into FFN.
\begin{equation}
    %\begin{small}
        \begin{aligned}
            H^{l+1}&=\text{ReLU}\Big(\big[A^{1}_M\big]\Big)W_{H}=\text{ReLU}\big(H^{l}W^{1}_{V}\big)W_{H}
        \end{aligned}
    %\end{small}
\end{equation}

In summary, SAN and FFN are two special cases in MANs with different static mask matrices. 

\begin{figure}[t]
    \centering
    \includegraphics[width=0.5\textwidth]{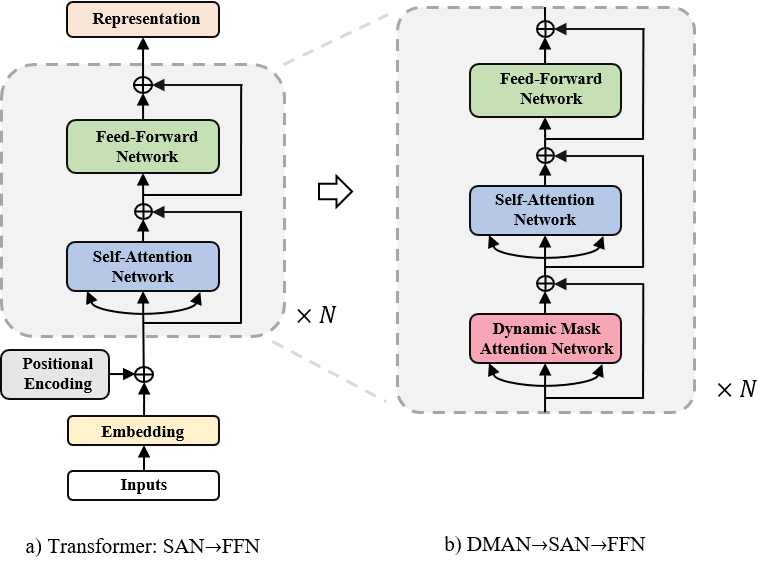}
	\caption{Overview of our proposed model. Left is the Transformer architecture, right is our DMAN$\to$SAN$\to$FFN one.}\label{model.figure}
\end{figure}

\subsection{Deficiency of SAN and FFN in Localness Modeling}
\label{deficiency-of-san-and-ffn-in-localness-modeling}
The mask matrix of SAN is an all-ones matrix and that of FFN is an identity matrix, they are two extreme cases in MANs. We analyze that these two static MANs are deficient in localness modeling.
Intuitively, through blocking other tokens in advance, FFN focuses on its own information and is unable to perceive the information except itself, let alone its neighbors. In SAN, each token is equally accessible to any other ones. As the example in Introduction shows, we find that tokens not in neighborhood are also likely to attend to each other with relatively large scores. Therefore, SAN might introduce noises to semantic modeling and overlook the relation of neighboring signals.

We demonstrate the issue of self-attention. Generally assuming that $\big[a,b,c\big]$ appear in sequence, and $(a, b), (b, c)$ are two neighbor pairs, but $a, c$ are not neighbors. % and therefore semantically irrelevant. 

First, to explicitly define the relationship of tokens, we introduce ${U_{\delta}(h)}$ as the set of tokens at the distance of $\delta$ from $h$ with key and query linear transformation in SAN, in other words, $u\in U_{\delta}(h)\Leftrightarrow ||hW_{Q}-uW_{K}||_{2}^{2}\leq\delta$. For example, if $(a, b)$ is a neighbor pair, there would exist some small $\delta\geq 0$ such that $a\in{U_{\delta}(b)}$ and $b\in{U_{\delta}(a)}$.

% As we know that the larger the inner product is, the smaller the Euclidean distance is, and vice versa. We define ${U_{\delta}(h)}$ as the set of tokens at the distance of $\delta$ from $h$ with key and query linear transformation, $u\in U_{\delta}(h)\Leftrightarrow ||hW_{Q}-uW_{K}||_{2}^{2}\leq\delta$. 

Second, we know that the larger the inner product is, the smaller the Euclidean distance is, and vice versa. With the awareness of the relationships between $\big[a,b,c\big]$, we have $a,b\in{U_{\delta}(a)}$, $b,c\in{U_{\delta}(c)}$ and $a,b,c\in{U_{\delta}(b)}$ for some small $\delta\geq 0$. %we can find some $\delat$ that $a\notin{U_{9\delta}(c)}$ and  $c\notin{U_{9\delta}(a)}$.

%Assuming that $\big[a,b,c\big]$ appear in sequence. $a\notin{U_{9\delta}(c)}$ and 
%$c\notin{U_{9\delta}(a)}$, which means that $a,c$ is not beneficial for each other's local semantic modeling. Moreover, $a,b\in{U_{\delta}(a)}$, $b,c\in{U_{\delta}(c)}$ and $a,b,c\in{U_{\delta}(b)}$. % We can see that 
Third, we are able to estimate the semantic distance between $a$ and $c$ as the Equation~\ref{estimation} shows.
\begin{equation}
    %\begin{small}
        \begin{aligned}
            &||aW_{Q}-cW_{K}||_{2}^{2}\\
            =&||aW_{Q}-bW_{K}+bW_{K}-bW_{Q}+bW_{Q}-cW_{K}||_{2}^{2} \\
            \leq &3||aW_{Q}-bW_{K}||_{2}^{2}+3||bW_{K}-bW_{Q}||_{2}^{2} \\+&3||bW_{Q}-cW_{K}||_{2}^{2}\big)\leq 9\delta
        \end{aligned}
    %\end{small}
    \label{estimation}
\end{equation}
Thus, though $a$ and $c$ are not neighbors, no matter how irrelevant the semantics of $a$ and $c$, $c\in {U_{9\delta}(a)}$ that $c$ would play an important role in modeling semantics of $a$. 

% which means that the inner product of $a$ and $c$ in SAN would be large and $c$ takes an important position in localness modeling of $a$, leads to a confliction. 

The upper phenomenon illustrates following normal attention function in Equation~\ref{self-attention}, some tokens not in neighborhood not are still likely to occupy an important position in attention weight that can not be ignored. % We conclude that in order to model the local structure, each token $h_{t}^{l}$ should assign more weights to its neighbors beyond the normal self-attention mechanism. 

\subsection{Dynamic Mask Attention Network}
\label{dynamic-mask-self-attention-network}
% Recall that SAN is expert in globalness modeling and FFN focuses on self-evolution, we argue that the intermediate state of these two mask matrices would be more suitable for localness modeling.

%the distance between $A$ and $C$ is impossible to be large. Quite a part of information in $C$ would also be absorbed into $A$ following attention mechanism, which would do harm to the local structure modeling of $A$. This sample illustrates that SAN ineffectively model the local structure.
%We assume that they are the boundary of $\big\{\mathcal{M}\big\}$ and the power of their collaboration has been verified. we hope other mask functions in $\big\{\mathcal{M}\big\}$ is able to join in and strengthen the collaboration.  

%In order to model the local structure in sentence, it is necessary that the particular token focuses on its own neighborhood, which means that the distance between the target token and those tokens outside the neighborhood should be large.

%Based on Masked Attention Network Family, we propose Dynamic Mask Self-Attention Network (DMAN). In the network, we add a prior position-dependent scalar to inform the attention function the relative distance between the query token and the target token with a more explicit manner, and the corresponding similarity scores can multiply these position-dependent scalars to obtain a new weight. The process is listed below:

With the knowledge of MANs, we propose to mask other tokens that not in neighborhood of the target token for better local semantic modeling.  

% Based on Mask Attention Networks, we build a distance-dependent mask matrix. It is able to inform the mask attention function of the relative distance between the query token and the key token in an explicit manner, then the corresponding attention weights can multiply these distance-dependent scalars to obtain new weights. 
%The process is listed below:
%\begin{gather}
%    \mathcal{S}_{M}(Q,K)=\Bigg[\frac{m(i-j)\text{exp}\big(Q_{i}K_{j}^{T}/\sqrt{d_{k}}\big)}{\sum_{k}m(i-k)\text{exp}\big(Q_{i}K_{k}^{T}/\sqrt{d_{k}}\big)}\Bigg]
%\end{gather}

% For example, we build a distance-dependent mask matrix $M^{S}$. It is able to inform the mask attention function of the relative distance between the query token and the key token in an explicit manner, then the corresponding attention weights can multiply these distance-dependent scalars to obtain new weights. 
For example, we build a distance-dependent mask matrix \text{SM}. If each token only model the relationship with those tokens within $b$ units of itself, we can set 
\begin{equation}
    %\begin{small}
        \begin{aligned}
            \text{SM}[t,s]=\left\{\begin{array}{cc}
             0, & |\ t-s\ |> b \\
             1, & |\ t-s\ |\le b
            \end{array}\right. \label{static-mask}
        \end{aligned}
    %\end{small}
\end{equation}
where $t,s$ are the positions of query and key, and $\text{SM}[t,s]$ is the value of the $t$-th row and $s$-th column of $\text{SM}$ . 

By means of \text{SM}, we take those tokens within $b$ units into account and ignore others. The static mask does assign more weights to a specific neighborhood, but lacks flexibility. Considering the neighborhood size varies with different query tokens, number of tokens that benefit for different query tokens' local semantic representation are different. Moreover, their mask matrices should match different attention heads and layers in MANs.

We propose Dynamic Mask Attention Network (DMAN) that replaces the static mask matrix. Incorporating query tokens, relative distance, attention head and layer, we build a dynamic mask function which replaces the hard $0/1$ mask gate in Equation~\ref{static-mask} with a soft one through sigmoid activation function in Equation~\ref{dynamic-mask-matrix}.
\begin{equation}
    %\begin{small}
        \begin{aligned}
    \text{DM}_{i}^{l}[t,s]=\sigma\Big(h^{l}_{t}W^{l}+P^{l}_{t-s}+U^{l}_{i}\Big)~\label{dynamic-mask-matrix}
        \end{aligned}
    %\end{small}
\end{equation}
where $s,t$ are the positions of query and key, $i$ is the attention head, $l$ is the layer. ${P}^{l}_{t-s}$ is parameterized scalar for the positions $t$ and $s$, $U^{l}_{i}$ is for the $i$-th head, and $W^{l}\in\mathbb{R}^{d\times 1}$. $W^{l}$, ${P}^{l}_{t-s}$ and $U^{l}_{i}$ are trainable parameters.

\begin{comment}
Next, we show that the mask matrix in Equation~\ref{dynamic-mask-matrix} is the intermediate state of SAN and FFN. Let $P^{l}_{t-s}\to{\infty}$ and $P^{l}_{t-s}\to{-\infty},t\neq s$, respectively. We can easily infer that for fixed $U^{l}_{i}$ and $W^{l}$:
\begin{equation}
   %\begin{small}
        \begin{gathered}
    \mathop{\lim}_{P^{l}_{t-s}\to{+\infty}}M^{D}_{t,s,i,l}=1,\ \mathop{\lim}_{P^{l}_{t-s}\to{-\infty},\ t\neq s}M^{D}_{t,s,i,l}=0~\label{lim}
        \end{gathered}
    %\end{small}
\end{equation}

The result demonstrates that the two extreme states of $M^{D}$ have the same functions with $[1]$ and $\mathbb{I}$, respectively. In other words, SAN and FFN are different extreme states of DMAN. Intuitively, if we set the capacity of neighborhood small enough, DMAN degenerates into FFN; if the capacity is large enough, DMAN changes into SAN. 
\end{comment}

\begin{table*}
\begin{center}
% \small
\begin{tabular}{lcccccc}
\midrule[1.0pt]
\multirow{2}{*}{Model} &\multicolumn{2}{c}{IWSLT14 De-En} &\multicolumn{4}{c}{WMT14 En-De}\\
%\cline{2-4}
&small &params &base &params &big &params \\
\midrule[1.0pt]
Transformer~\cite{vaswani2017attention} &34.4 &36M &27.3 &62M &28.4 &213M \\
%\hline
Convolutional Transformer~\cite{yang2019convolutional} &- &- &28.2 &88M &28.7 &-  \\
Weighted Transformer~\cite{ahmed2017weighted} &- &- &28.4 &65M &28.9 &213M \\
Local Transformer~\cite{yang2018modeling} &- &- &28.5 &89M &29.2 &268M \\
%\hline
Relative Transformer~\cite{shaw2018self} &- &- &26.8 &- &29.2 &- \\
%\hline
%Universal Transformer~\cite{dehghani2018universal} &- &28.9 &- \\
%\hline
Scaling NMT~\cite{ott2018scaling} &- &- &- &- &29.3 &213M \\
%\hline
Dynamic Conv~\cite{wu2019pay} &35.2 &- &- &- &29.7 &213M \\
% MultiScale Col~\cite{} &- &- &- &- &
%\hline
% Macaron Net~\cite{lu2019understanding} &35.4 &36M &28.9 &62M &30.2 &213M \\
\midrule[1.0pt]
%\hline
%\hline
Ours &\textbf{36.3} &37M & \textbf{29.1} &63M & \textbf{30.4} &215M \\
\midrule[1.0pt]
%\hline
\end{tabular}
\end{center}
\caption{Translation performance (BLEU) on IWSLT14 De-En and WMT14 En-De testsets.}
\label{translation-result}
\end{table*}

%\subsection{The Advantage of Dynamic Mask Self-Attention Network}
\subsection{Collaboration of Mask Attention Networks}
\label{collaboration-of-mask-attention-network-family}
%Note that Self-Attention Network (SAN) does not mask any tokens and specializes in long-range dependency, Feed-Forward Network (FFN) masks all tokens except itself that focuses on self-processing, 
Until here, we have three sub-networks of MANs, namely, SAN, FFN and DMAN. SAN that does not mask any tokens and specializes in global semantic modeling. FFN that masks all tokens except itself and focuses on self-processing. DMAN masks the tokens not in neighborhood and is able to model local structure more effectively. 

Transformer is composed of SAN and FFN that achieves positive results in various NLP tasks, the stacking method of Transformer inspires us to stack DMAN, SAN and FFN to incorporate their advantages. We insert DMAN in the manner of DMAN$\to$SAN$\to$FFN, which is shown in Figure~\ref{model.figure}. With this architecture, we first model the localness then globalness, and take the step for self-evolution in the end.

\section{Experiments}
In this section, we introduce our experiments. We first describe the experimental details in \S~\ref{experiment-details}. Then we show the experimental results in \S~\ref{experiment-results}. Finally we conduct the ablation study and analysis in \S~\ref{ablation-study}.

\subsection{Experimental Setting}
\label{experiment-details}
% The implementation details would be described in the appendix.
\subsubsection{Machine Translation}
Machine translation is an important application of natural language processing~\cite{vaswani2017attention}. We evaluate our methods on two widely used public datasets: IWSLT14 German-to-English (De-En) and WMT14 English-to-German (En-De). IWSLT14 De-En dataset consists of about 153K/7K/7K sentence pairs for training/validation/testing. WMT14 En-De dataset consists of about 4.5M sentence pairs, and the models were validated on newstest2013 and examined on newstest2014.

Our data processing follows~\citet{lu2019understanding}. For IWSLT2014, we set our model into the small one, the hidden size, embeddings and attention heads to 512, 512, and 4 respectively. For the WMT14 dataset, following the Transformer setting of~\citet{vaswani2017attention}, we set our model into the base and big ones which both consist of a 6-layer encoder and 6-layer decoder, the hidden nodes are set to 512 and 1024, and the number of attention heads are 8 and 16. For each setting (small, base and big), we replace all layers in Transformer by our MAN layer. To make a relatively fair comparison, we set the dimensionality of the inner-layer of the FFN in the MAN layers to two times of the dimensionality of the hidden states.

We train our proposed model with cross-entropy with 0.1 label smoothing rate. Inverse-sqrt learning rate scheduler are employed, the peak learning rates are 1.5e-2, 1e-2 and 7e-3 with 8k warmup, 50k update, 80k update and 80k update for transformer big, base and small model with max-tokens 4096, 12288 and 8192 per batch. The dropout rates are 0.3, 0.1 and 0.3 for small, base and big models. The optimizer of model is Adam with (0.9,0.98). The beam size and length penalty for base and big models are 4 and 0.6, for small model is 5 and 1.0. The base and large model are trained on 8 V100 GPUs, and the small model is trained on 2 P40.

\subsubsection{Abstract Summarization}
\label{abstract-summarization}
Automatic summarization aims to produce a concise and fluent summary conveying the key information in the input text. We focus on abstractive summarization, a generation task where the summary is not limited in reusing the phrases or sentences in the input text. We use the CNN/Daily Mail~\cite{see2017get} and Gigaword~\cite{rush2015neural} for model evaluation. 

Following \citet{song2019mass}, we set the hidden size, embeddings and attention heads to 768, 768, and 12 respectively. Our model consists of a 6-layer encoder and 6-layer decoder. For the convenience of comparison, the training follows classic seq2seq model without copy, converge or RL. We remove duplicated trigrams in beam search~\cite{paulus2017deep}. Moreover, the dimensionality of the inner-layer of the FFN in the MAN layers is set to two times of the dimensionality of the hidden states.

In training, inverse-sqrt learning rate scheduler is employed. The peak learning rates are 1e-3 and 8e-4, max-tokens per batch are 8192 and 12288 for CNN/Daily Mail and Gigaword, respectively. The warmup steps is 8k and the total updates is 50k. The optimizer of model is Adam with (0.9,0.98). The dropout and clip-norm are both 0.1. During decoding, the beam size are both 5, the max length and length penalty are 50 and 2.0 for CNN/Daily Mail, 30 and 1.0 for Gigaword. The models are trained on 4 P40 GPUs.

\begin{table*}
\begin{center}
%\small
\begin{tabular}{lcccccccc}
\midrule[1.0pt]
\multirow{2}{*}{Model} &\multicolumn{4}{c}{CNN/Daily Mail} &\multicolumn{4}{c}{Gigaword}\\
&R-1 &R-2 &R-L &R-avg &R-1 &R-2 &R-L &R-avg \\
\midrule[1.0pt]
LEAD-3~\cite{nallapati2016abstractive} 
&40.42 &17.62 &36.67 &31.57 &- &- &- &-\\
PTGEN+Coverage~\cite{see2017get} 
&39.53 &17.28 &36.38 &31.06 &- &- &- &- \\
Dynamic Conv~\cite{wu2019pay} 
&39.84 &16.25 &36.73 &30.94 &- &- &- &- \\
Transformer~\cite{vaswani2017attention} 
&39.50 &16.06 &36.63 &30.73 &37.57 &18.90 &34.69 &30.38 \\
\midrule[0.5pt]
Ours &\textbf{40.98} &\textbf{18.29} &\textbf{37.88} &\textbf{32.38} &\textbf{38.28} &\textbf{19.46} &\textbf{35.46} & \textbf{31.06} \\
\midrule[1.0pt]
\end{tabular}
\end{center}
\caption{Evaluation results on CNN/Daily Mail and Gigaword. R is short for ROUGE.}
\label{summarization-result}
\end{table*}

\subsection{Experimental Results}
\label{experiment-results}
\subsubsection{Machine Translation}
In machine translation, BLEU~\cite{papineni2002bleu} is employed as the evaluation measure. Following common practice, we use tokenized case-sensitive BLEU and case-insensitive BLEU for WMT14 En-De and IWSLT14 De-En, respectively. We take Transformer~\cite{vaswani2017attention} as the baseline and compare with other concurrent methods. Convolutional Transformer~\cite{yang2019convolutional} restricts the attention scope to a window of neighboring elements in order to model locality for self-attention model. 
% Weighted Transformer~\cite{ahmed2017weighted} replaces the multi-head attention by multiple self-attention branches that the model learns to combine during the training process.
Local Transformer~\cite{yang2018modeling} casts localness modeling as a learnable Gaussian bias, which indicates the central and scope of the local region to be paid more attention. 
%Relative Transformer~\cite{shaw2018self} extends the self-attention mechanism to efficiently consider representations of the relative distances between sequence elements.
% Dynamic Conv~\cite{wu2019pay} introduces dynamic convolutions to predict separate convolution kernels solely based on the current time-step in order to determine the importance of context elements. 
%Macaron Net~\cite{lu2019understanding} adds one more FFN sublayer in the front of transformer layer based on Strang-Marchuk splitting scheme in ODE.

The results for machine translation are shown in Table~\ref{translation-result}.
Our model exceeds the baseline Transformer and other models. For the IWSLT14 dataset, our small model outperforms the Transformer small by 1.6 points in terms of BLEU. For the WMT14 dataset, our base model exceeds its Transformer counterpart by 1.8 BLEU points. Furthermore, the performance of our base model is even better than that of the Transformer big model reported in~\cite{vaswani2017attention}, but with much less parameters. Our big model outperforms the Transformer big by 2.0 BLEU points. 

Compare with Convolutional Transformer and Local Transformer, our model also achieve 1.7 and 1.2 points improvement in BLEU, respectively. This validates that the superiority of our model to systematically solve the localness modeling problem in Transformer.

\subsubsection{Abstractive Summarization}
We use the F1 score of ROUGE~\cite{lin2003automatic} as the evaluation metric\footnote{\url{https://github.com/pltrdy/files2rouge}}. In Table~\ref{summarization-result}, we compare our model against the baseline Transformer~\cite{vaswani2017attention} and several generation models on CNN/Daily Mail and Gigaword. LEAD3~\cite{nallapati2016abstractive} extracts the first three sentences in a document as its summary. PTGEN+Converage~\cite{see2017get} is a sequence-to-sequence model based on the pointer-generator network. As shown in Table~\ref{summarization-result}, our model outperforms Transformer by 1.4 in ROUGE-1, 2.2 in ROUGE-2 and 1.2 in ROUGE-L in CNN/Daily Mail. In Gigaword dataset, ours exceeds the baseline by 0.7 in ROUGE-1, 0.5 in ROUGE-2 and 0.7 in ROUGE-L.  

As a summary, in machine translation and abstractive summarization our proposed model achieves better results than the Original Transformer~\cite{vaswani2017attention}.

\section{Further Analysis}
In this section, we conduct further analysis for our model. We first investigate stacking methods for different sublayers in \S~\ref{Investigate_Stacking_Methods}. Then we compare strategies of static mask and dynamic mask in \S~\ref{Static_Mask_and_Dynamic_Mask}. Finally, we analyse the behavior of SAN and DMAN in localness modeling through attention scores in \S~\ref{Analysis_of_DMAN_in_Localness_Modeling}.
\label{ablation-study}
\subsection{Investigate Stacking Methods for Different Sublayers}
\label{Investigate_Stacking_Methods}
Here, we investigate different collaboration mechanisms of the elements in MANs. Under our design principles, there are three elements: FFN, SAN, and DMAN. For the convenience of comparison, we take FFN as the last component in the sequential layered structure. We try different collaboration methods and test them on IWSLT2014 German-to-English (De-En). The results are shown in the Table~\ref{ablation-result}. We conclude that:

\begin{table}
\begin{center}
%\small
\begin{tabular}{clc}
\midrule[1.0pt]
\# &Method &BLEU \\
\midrule[1.0pt]
$C$\#1 & FFN$\to$SAN$\to$FFN &35.51 \\
$C$\#2 & SAN$\to$SAN$\to$FFN &35.66 \\
$C$\#3 & DMAN$\to$DMAN$\to$FFN &35.86 \\
$C$\#4 & SAN$\to$DMAN$\to$FFN &35.91 \\
\midrule[1.0pt]
$C$\#5 & DMAN$\to$SAN$\to$FFN  & \textbf{36.35} \\
\midrule[1.0pt]
\end{tabular}
\end{center}
\caption{Performance of different collaboration methods of DMAN, SAN and FFN. We evaluate on IWSLT2014 De-En.}
\label{ablation-result}
\end{table}

\begin{enumerate}
    \item Our proposed  $C$\#5 achieves the best performance that verify the effectiveness of our proposed sequential layered structure.
    \item All of $C$\#3, $C$\#4 and $C$\#5 outperform $C$\#1 and $C$\#2, and the least improvement in BLEU is 0.2. This shows that no matter what collaboration method, models with the participation of DMAN perform better than models without DMAN, which validates the capability of DMAN.
    \item Both $C$\#5 and $C$\#4 are better than $C$\#3 and $C$\#2. This indicates that models without DMAN or SAN are not comparable to models with all three modules. This shows that DMAN and SAN have their own strengths, namely, localness modeling and globalness modeling, and are able to make up for each other's defects through collaboration.
    \item $C$\#5 is better than $C$\#4.
    This indicates that first modeling the localness and then globalness would be better than the inverse order.  
\end{enumerate}

\subsection{Static Mask and Dynamic Mask}
\label{Static_Mask_and_Dynamic_Mask}
In this section, we compare the performance of Static Mask Attention Network (SMAN) and Dynamic Mask Attention Network (DMAN). Both of them follow the collaboration strategy of DMAN(SMAN)$\to$SAN$\to$FFN. 
In SMAN, we set a fixed mask boundary which has been determined in advance following Equation~\ref{static-mask}. Empirically, we propose two static mask strategies: (a) SMAN$_{1}$, the boundary $b$ depends on sentence length $L$, $b=\sqrt{L}/2$; (b) SMAN$_{2}$, $b$ is set to 4, which is chosen from 2, 4, 6, 8 through validation. %, regardless of any other external factors.

The results in IWSLT2014 De-En are shown in Table~\ref{static-ablation-result}. The performance of SMAN$_{1}$ and SMAN$_{2}$ are very close. They both outperform the Transformer but fall behind our proposed DMAN. 
This indicates that our proposed DMAN is superior to SMAN. SMAN fails to manage various neighborhood for different query tokens, but DMAN can model localness with more flexibility according to these factors.

\begin{table}
\begin{center}
% \small
\begin{tabular}{lc}
%\hline
\midrule[1.0pt]
% &Transformer &SMAN$_{1}$ &SMAN$_{2}$ &DMAN \\
model &BLEU \\
\midrule[1.0pt]
%BLEU &34.40 &35.52 &35.55 &\textbf{36.05} \\
Transformer  &34.40 \\
SMAN$_{1}$ &35.52 \\
SMAN$_{2}$ &35.55 \\
DMAN &\textbf{36.35}\\
\midrule[1.0pt]
%\hline
\end{tabular}
\end{center}
\caption{Performance of SMAN and DMAN on IWSLT2014 De-En.}
\label{static-ablation-result}
\end{table}

\subsection{Analysis of DMAN in Localness Modeling}
\label{Analysis_of_DMAN_in_Localness_Modeling}
In this section, we analyse the behavior of DMAN and SAN in localness modeling through attention scores in Equation~\ref{mask_attention_function_1}. To quantify the role of neighbors in semantic modeling, we compute the sum of attention scores within some particular window size.  Generally, if the attention score from $a$ to $c$ is bigger than $b$ to $c$, we consider that $a$ contributes more to the semantic modeling of $c$ compared to $b$, in other words, 
model utilizes more information of $a$ than $b$ to learn the semantic representation of $c$. Therefore, larger attention scores mean that model utilizes more information of the corresponding tokens to learn the semantic representation of query token. 

For each sentence in dataset $X_{i}=(x_{i,1},\cdots,$ $x_{i,T_{i}})\in\mathcal{D}$, we utilize $\bar{s}^{l}_{i,\emph{DMAN}}$ and $\bar{s}^{l}_{i,\emph{SAN}}\in\mathbb{R}^{T_{i}\times T_{i}}$ to denote the average attention scores $\mathcal{S}_{M}(Q,K)$ in Equation~\ref{mask_attention_function_1} across different heads in the $l$-th layer for DMAN and SAN, respectively. We sum the attention scores of these tokens $x_{i,k}$ within the window size $w$ of the query $x_{i,j}$ in the $l$-th layer, and average the sum across $X_{i}$ and dataset $\mathcal{D}$ following Equation~\ref{indicator}.
\begin{equation}
    %\begin{small}
        \begin{gathered}
        attn\_s_{w,l,*}=\frac{1}{|\mathcal{D}|}\sum_{X_{i}\in\mathcal{D}}\frac{1}{T_{i}}\sum_{x_{i,j}\in X_{i}}\sum_{|k-j|\le w}\bar{s}^{l}_{i,*}\big[j,k\big]~\label{indicator}
        \end{gathered}
    %\end{small}
\end{equation}
where $*\in\{\emph{DMAN},\emph{SAN}\}$, and $\bar{s}^{l}_{i,*}\big[j,k\big]$ is the value of the $j$-th row and $k$-th column of $\bar{s}^{l}_{i,*}$. $attn\_s_{w,l,*}$ measures the overall contribution of these neighbor tokens within the window size $w$ to the query tokens' semantic modeling. We take $\mathcal{D}$ as the test set of IWSLT14 De-En and compute $attn\_s_{w,l,*}$ with $w=1,2,4$ and $l=1,3,6$. 

\begin{table}
\begin{center}
% \small
\begin{tabular}{l|c|ccc}
%\hline
\midrule[1.0pt]
	&$w$ &\#1 &\#3	&\#6 \\
\midrule[1.0pt]
\emph{DMAN}	&1	&76.58	&60.43	&60.86 \\
\emph{SAN}	&1  &12.80	&40.39	&45.55 \\
\midrule[0.5pt]
\emph{DMAN}	&2  &86.17	&75.56	&73.89 \\
\emph{SAN}	&2  &18.73	&45.62	&52.72 \\
\midrule[0.5pt]
\emph{DMAN}	&4  &95.09	&86.20	&85.58 \\
\emph{SAN}	&4  &30.38	&55.17	&62.77 \\
\midrule[1.0pt]
%\hline
\end{tabular}
\end{center}
\caption{The values of attention scores $attn\_s_{w,l,\emph{DMAN}}$ and $attn\_s_{w,l,\emph{SAN}}$, which is shown in Equation~\ref{indicator}. $\mathcal{D}$ is the test set of IWSLT14 De-En, window size $w=1,2,4$ and encoder layers $l=1,3,6$. }
\label{localnessDMAN}
\end{table}

 The result is shown in Table~\ref{localnessDMAN}. We see that in layer\#1, \#3 and \#6, the sum attention scores of DMAN within the window size $2$ are 50\% more than those of SAN, especially in layer\#1 where the gap is as much as five times between SAN and DMAN. This phenomenon validates that the attention scores of DMAN in neighbors are larger than those of SAN, thus DMAN is more specialized in localness modeling than SAN. 

\section{Related Work}
Recently, there is a large body of work on improving Transformer~\cite{vaswani2017attention} for various issues. 
For recurrence modeling,~\citet{hao2019modeling} introduces a novel attentive recurrent network to leverage the strengths of both attention and recurrent networks.
For context modeling,~\citet{yang2019context} focuses on improving self-attention through capturing the richness of context and proposes to contextualize the transformations of the query and key layers.
\citet{wu2019pay} introduces dynamic convolutions to predict separate convolution kernels solely based on the current time-step in order to determine the importance of context elements.
In order to adjust attention weights beyond SAN, \citet{shaw2018self} extends the self-attention mechanism to efficiently consider representations of the relative positions or distances between sequence elements through adding a relative position embedding to the key vectors;
\citet{bugliarello2019improving} transfers the distance between two nodes in dependency trees with a pre-defined Gaussian weighting function and multiply the distance with the key-query inner product value; %presents a novel method to incorporate syntactic information in the self-attention mechanism of the Transformer encoder by introducing attention heads that can attend to the dependency parent of each token. They 
\citet{dai2019transformer} presents a relative position encoding scheme that adds additional relative position representation to the key-query computation.
\citet{sukhbaatar2019adaptive} proposes a parameterized linear function over self-attention to learn the optimal attention span in order to extend significantly the maximum context size used in Transformer. 
To merge FFN to SAN,~\citet{sukhbaatar2019augmenting} proposes a new model that solely consists of attention layers and augments the self-attention layer with persistent memory vectors that play a similar role as the feed-forward layer.
As for the collaboration of SAN and FFN, \citet{lu2019understanding} introduces Macaron layer that split the FFN into two half-steps based on Strang-Marchuk splitting scheme in ODE.
For localness modeling,~\citet{yang2018modeling} casts localness modeling as a learnable Gaussian bias according to relative distance to external energy in softmax function as a new self-attention network.~\citet{zhao2019muse} explores parallel multi-scale representation learning to capture both long-range and short-range language structures with combination of convolution and self-attention.
In our work, DMAN, SAN and FFN are unified in Mask Attention Networks, where DMAN is a supplement of SAN and FFN that specializes in localness modeling.  Moreover, we investigate different collaboration mechanisms. 

\section{Conclusion}
In this paper, we introduce Mask Attention Networks and reformulate SAN and FFN to point out they are two special cases with static mask in MANs. We analyze the %advantage of SAN in long-range dependency modeling and that of FFN in self-evolution, and point out their drawbacks 
the deficiency of SAN and FFN in localness modeling. Dynamic Mask Attention Network is derived from MANs for better local structure modeling. Considering the different specialities of SAN, FFN, and DMAN, we investigate a sequential layered structure DMAN$\to$SAN$\to$FFN for their collaboration. Compared with original Transformer, our proposed model achieves better performance in neural machine translation and abstract summarization. For future work, we consider adding structure information or external knowledge, e.g., dependency tree, with mask matrices in MANs.

\section{Acknowledgement}
This work is partially supported by National Natural Science Foundation of China (No.71991471), Science and Technology Commission of Shanghai Municipality Grant (No.20dz1200600).

\bibliography{anthology,custom,naacl2021}
\bibliographystyle{acl_natbib}

\end{document}